\documentclass[9pt,shortpaper,twoside,web]{ieeecolor2}
\usepackage{generic}
\usepackage{orcidlink}
\usepackage{cite}
\usepackage{amsmath,amssymb,amsfonts}
\usepackage{algorithmic}
\usepackage{graphicx}
\usepackage{textcomp}
\usepackage{booktabs}
\usepackage{graphicx}  
\usepackage{makecell}  
\def\BibTeX{{\rm B\kern-.05em{\sc i\kern-.025em b}\kern-.08em
    T\kern-.1667em\lower.7ex\hbox{E}\kern-.125emX}}
\begin{document}
\title{Monocular Markerless Motion Capture Enables Quantitative Assessment of Upper Extremity Reachable Workspace }
\author{
  Seth Donahue, J.D. Peiffer, R. Tyler Richardson, Yishan Zhong, 
  Shaun Q. Y. Tan, \textit{Graduate Student Member, IEEE}, Benoit Marteau, \textit{Graduate Student Member, IEEE}, 
  Stephanie R. Russo, May D. Wang, \textit{Fellow, IEEE}, R. James Cotton, Ross Chafetz
\thanks{This work has been submitted to the IEEE for possible publication. Copyright may be transferred without notice, after which this version may no longer be accessible.}
\thanks{S. Donahue \orcidlink{0000-0002-8387-9887} is with Shriners Children's Lexington, Lexington, KY 40508, USA and University of Kentucky Department of Physical Therapy, Lexington, KY 40536 (e-mail: seth.donahue@shrinenet.org).}
\thanks{R. J. Cotton \orcidlink{0000-0001-5714-1400} is with Shirley Ryan AbilityLab, Chicago, IL 60611, USA and Northwestern University, Evanston, IL 60208, USA (e-mail: rcotton@sralab.org).}
\thanks{J.D. Peiffer \orcidlink{0000-0003-2382-8065} is with the Shirley Ryan Ability Lab, Center for Bionic Medicine, Chicago, IL 60611, USA and Northwestern University, Evanston, IL 60208, USA (e-mail: jpeiffer@sralab.org).}
\thanks{S. Tan, Y. Zhong, and B. Marteau are with School of Electrical and Computer Engineering, Georgia Institute of Technology, Atlanta, GA 30332, USA (e-mail: stan99@gatech.edu; yzhong307@gatech.edu; benoitmarteau@gatech.edu).}
\thanks{M. D. Wang is with The Wallace H. Coulter Department of Biomedical Engineering, Georgia Institute of Technology and Emory University, Atlanta, GA 30322, USA (phone: 404-385-2954; e-mail: maywang@gatech.edu).}
\thanks{R. Chafetz is with Shriners Hospitals for Children, Philadelphia, PA 19140, USA (e-mail: rchafetz@shrinenet.org).}
\thanks{R. T. Richardson \orcidlink{0000-0001-6449-6613} is with Pennsylvania State University at Harrisburg, Middletown, PA 17057, USA (e-mail: rtr12@psu.edu).}
\thanks{S. Russo is with Nationwide Children's Hospital, Columbus, OH 43205, USA (e-mail: sarusso@udel.edu).}
}

\maketitle
\begin{abstract}
\textbf{Objective:} To validate a clinically accessible approach for quantifying the Upper Extremity Reachable Workspace (UERW) using a single (monocular) camera and Artificial Intelligence (AI)-driven Markerless Motion Capture (MMC) for biomechanical analysis. Objective assessment and validation of these techniques for specific clinically oriented tasks are crucial for their adoption in clinical motion analysis.  AI-driven monocular MMC reduces the barriers to adoption in the clinic and has the potential to reduce the overhead for analysis of this common clinical assessment \textbf{Methods:} Nine adult participants with no impairments performed the standardized UERW task, which entails reaching targets distributed across a virtual sphere centered on the torso, with targets displayed in a VR headset. Movements were simultaneously captured using a marker-based motion capture system and a set of eight FLIR cameras. We performed monocular video analysis on two of these videos to compare a frontal and offset camera configuration. Agreement between the MMC and marker-based systems was evaluated by comparing the percentage of targets reached across six workspace octants, and the octant agreement between the systems.
\textbf{Results:} The frontal camera orientation demonstrated strong agreement with the marker-based reference, exhibiting a minimal mean bias of $0.61 \pm 0.12$ \% reachspace reached per octanct (mean $\pm$ standard deviation). In contrast, the offset camera view underestimated the percent workspace reached ($-5.66 \pm 0.45$ \% reachspace reached). Depth-related errors in the frontal configuration were primarily confined to posterior octants, whereas the offset view presented inaccuracies in both contralateral and posterior octants. \textbf{Conclusion:} The findings support the feasibility of a frontal monocular camera configuration for UERW assessment, particularly for anterior workspace evaluation where agreement with marker-based motion capture was highest. While posterior workspace accuracy remains limited depth-estimation and anatomical occlusion errors, the overall performance demonstrates clinical potential for practical, single-camera assessments. \textbf{Significance:} This study provides the first validation of monocular MMC system for the assessment of the UERW task. By reducing technical complexity and equipment demands, this approach enables broader implementation of quantitative upper extremity mobility assessment in both clinical and home-based rehabilitation settings.
\end{abstract}

\begin{IEEEkeywords}
Reachable Workspace, Markerless Motion Capture, Monocular Camera, Clinical Feasibility, Artificial Intelligence 
\end{IEEEkeywords}

\section{Introduction} 
Quantitative tracking of human movement is central to clinical assessment, enabling identification of biomechanical phenotypes that complement subjective evaluations and Patient-Reported Outcomes (PROs) \cite{GRAF2025, Lovette2024, richardson_reachable_2022, russo2019motion}. Traditionally, this tracking has been performed using marker-based motion capture, which is considered the gold standard for biomechanical analysis. Marker-based systems track anatomical landmarks using retro-reflective markers and have been widely used to assess Upper Extremity (UE) function\cite{russo_motion_2019,han2013validity, richardson_reachable_2022, russo2019motion}.

Over the past two decades, UE motion assessment has evolved from subjective scoring to objective quantification using marker-based systems \cite{abzug2010current, russo2019motion, koman2008quantification, pearl2014assessing}. However, many analyses have focused on a limited set of motions or static poses \cite{nicholson2014evaluating, russo2014scapulothoracic, russo2015limited}. In contrast, the UERW provides a holistic appraisal of global UE mobility by quantifying the regions in space that can be reached by the hand \cite{han2013validity}.  This iteration of the  UERW task involves patients reaching toward virtual targets displayed around them, providing a quantitative, biofeedback-driven measure of their reachable workspace \cite{richardson_reachable_2022}. Assessment of the UERW task is calculated by the number of targets reached compared to the number of targets available in the octant to reach and expressed as a percentage. Previous literature has established UERW as a valid, objective and clinically relevant measure to differentiate between patient populations and healthy controls, and to track rehabilitation progress over time \cite{richardson_reachable_2022, han_reachable_2015, han2015dystrophinopathy,de_bie_longitudinal_2017}. While highly accurate and clinically valuable, marker-based systems come with several practical challenges that make them difficult to implement routinely. This includes the need for specialized laboratory space, multiple cameras, careful calibration, and technical expertise for both setup and operation. In addition, patients often require significant preparation time, and data processing can be labor-intensive. Together, these factors make it challenging to integrate motion capture into everyday clinical workflows, limiting its accessibility for many patients and clinicians.

As an alternative to marker-based motion capture, MMC has emerged as a promising alternative. Initial work in this space included simple video-based methods for objective movement analysis using manual annotation of patients joints and segments, often performed frame by frame by an engineer or clinical team member \cite{Olleac2025VideoGait, Davids2006SHUEE, Kephart2020GaitApp}. However, these manual approaches are time-intensive and require both clinical and technical expertise. Advances in Artificial Intelligence (AI), computer vision, and human pose estimation now offer a pathway to overcome these limitations. Multi-camera systems, such as Theia3D, OpenCap, and open-source biomechanics frameworks, have demonstrated high accuracy joint location and angle estimation across diverse movement tasks\cite{wren_comparison_2023,hansen_validation_2024, kanko_inter-session_2021,uhlrich_opencap_2023, cotton2024differentiablebiomechanicsunlocksopportunities}. While these multi-camera MMC systems are generally accurate for UE assessment there are still limitations for their implementations in clinical settings due to their need for synchronized camera setups and calibration of the extrinsic parameters and overall complexity of the system \cite{unger_differentiable_2025, hansen_validation_2024,wade_applications_2022, guo2025artificial}.

Given the logistical challenges of multi-camera systems, single-camera (monocular) MMC offers a simpler, more scalable approach for point-of-care biomechanical reconstructions and analysis. Recent studies have shown that monocular biomechanical reconstruction with MMC performed comparably to marker-based and markerless reconstruction with high accuracy despite inherent challenges such as occlusion, depth estimation errors, and keypoint detection noise \cite{koleini2025biopose, CottonGaitTransformer}. While Monocular motion capture has the potential to be a powerful tool for clinical assessment and full body tracking, depth estimation is known to a potential issue with these systems in the measurement of clinically relevant variables \cite{pantzar2018knee, pierzchlewicz2024platyposecalibratedzeroshotmultihypothesis}.  Other data driven techniques that use  keypoints and keypoints combined with machine learning based outputs; these systems while accurate, do not have the necessary biomechanical modeling and clinical validity metrics associated yet to  inform clinical decision making \cite{Lagomarsino_2025, panattoni2024hand, klein2024angle, Tang_2025}. While these system have been shown to be clinical valid, they are purely data driven and lack biomechanical constraints that are necessary for clinical decision making. Recent work has combined an anatomically accurate full body tracking, human pose estimation, and biochemically accurate MuJoCo models in clinical populations with a single smartphone \cite{peiffer2025portable}. While this is an excellent tool that shows promise for the evaluation human movement at scale, it is yet be be validated on specific tasks or clinical assessments beyond gait analysis. To address this opportunity, the present study validates a single camera system for analyzing and scoring participants performing the UERW task with a single monocular camera based system.

To our knowledge, this is the first study to evaluate monocular MMC for the UERW task and to compare multi-camera configurations against the clinical gold standard (marker-based motion capture). We validate the use of a singe-camera biomechanical reconstruction for UERW from two independent camera configurations: frontal and offset. We hypothesize that the frontal configuration will outperform the offset view and find better agreement across each of the eight UERW octants due to the discrepancies in depth perception with the offset camera. The task selected for this work is significantly more challenging from a technical perspective as the UERW assessment is highly dynamic and consists of multi-planar movements, thus providing a robust assessment of current state of the art AI driven monocular systems.

\section{Methods}
\subsection{Participants}
Nine adults without movement impairments (5 males, 4 females; height: $1.69 \pm 0.09 \, \text{m}$, mass: $70.56 \pm 19.05 \, \text{kg}$) participated in this study. All participants were free of underlying conditions affecting upper extremity mobility and had no history of neurological injury within the past year. Only the UERW of the right arm was assessed, regardless of hand dominance.

\begin{figure*}[ht]
    \centering
    \includegraphics[width=\textwidth]{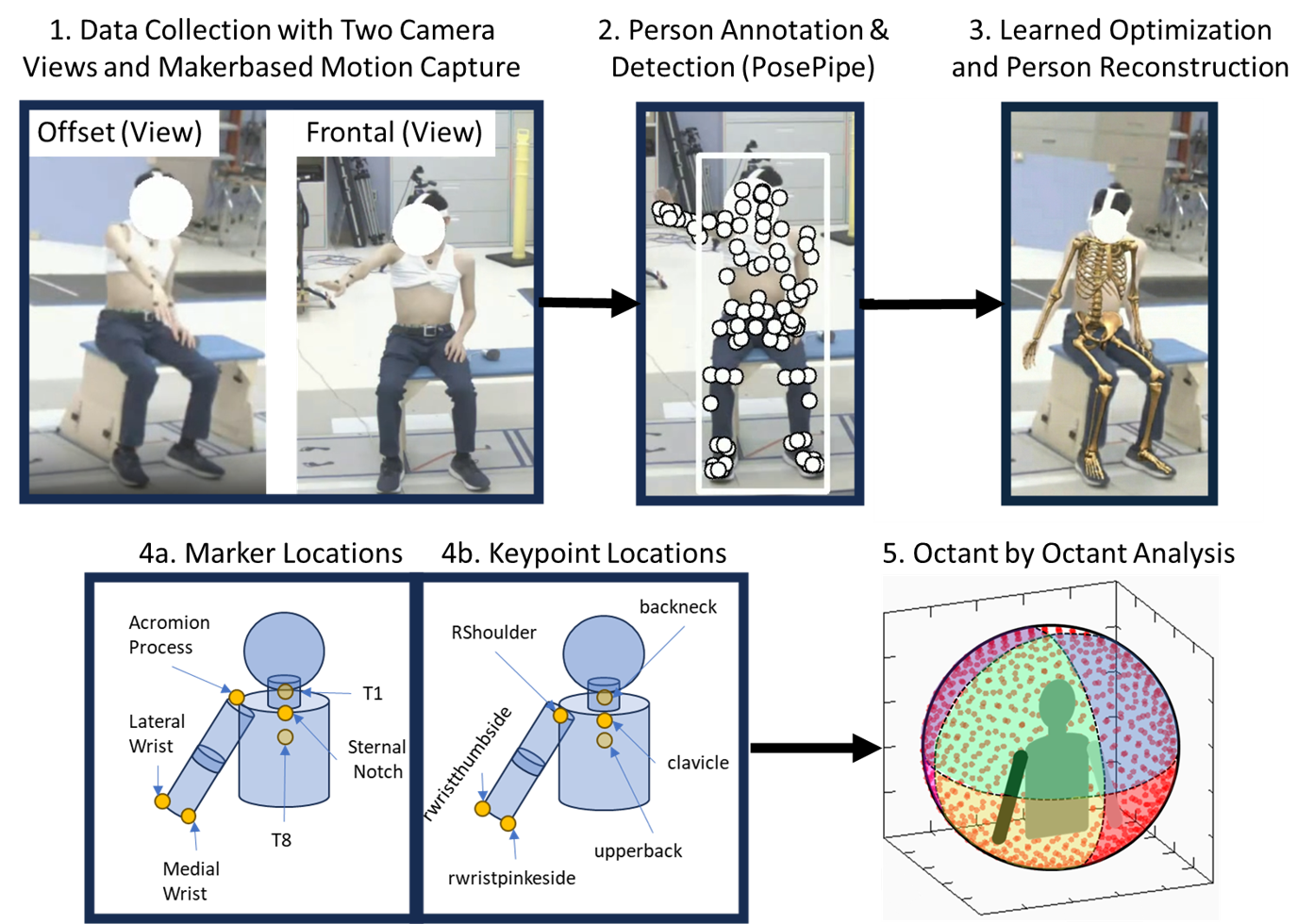}
    \caption{Methods overview. (1) Biomechanical reconstruction using two camera views: frontal and offset, compared to marker-based motion capture. (2) Video based processing included the steps descirbed in \cite{peiffer2025portable}, which included post processing from PosePipeline with the MeTRAbs Bottom Up Keypoint detection, EasyMoCap Person annotation.  (3) Presents the learned position optimization and biomechancial person reconstruction using MuJoCo physics engine . (4a) Marker locations placed by the same clinician for repeatability; (4b) keypoint positions from the MeTRABS detector used in analysis. (5) Workspace and octant-by-octant analysis of targets reached, with total available targets shown as red dots. Each shaded octant remains consistent throughout all figures.}
    \label{fig:methods_figure}
\end{figure*}

\subsection{Data Collection} \label{coordinate systems}
For each participant, data were collected simultaneously using a twelve camera marker-based motion capture system (Vicon Industries, Hauppauge, NY) and an eight camera video recording system (Blackfly S Forward-Looking Infrared or FLIR, Teledyne, Thousand Oaks, CA) through Vicon Nexus, sampled at 60 Hz. A gamified version of the UERW real-time feedback was presented using a Virtual Reality (VR) headset, the Meta Quest 3 (Meta, Menlo Park, CA), with pass-through vision \cite{vrraise}. Participants viewed an array of virtual targets, placed at arms length as estimated by the VR system, with 800 targets superimposed over the pass-through VR view and were instructed to reach for the targets with their right hand. When the hand moved within 5 cms of a virtual target, the target disappeared \cite{vrraise}. Participants continued reaching until all targets were achieved or no further targets could be reached \cite{richardson_evaluation_2024, vrraise}.

The analysis presented in this work is a post-hoc evaluation of two single views from the original eight FLIR video cameras analyzed independently. A frontal camera was positioned anterior and superior to the participant, while an offset view was positioned anterior, superior, and lateral, approximately $45^{\circ}$ to the right of the frontal camera (Fig.~\ref{fig:methods_figure}). Retroreflective markers were affixed to the skin of the participants using adhesive Velcro coins; all markers were placed by the same trained researcher to ensure consistency. Markers were affixed to the radial and ulnar styloids of the right wrist, the acromion process of the right scapula, the T1 and T8 spinous processes, and the sternal notch  (Fig.~\ref{fig:methods_figure}). A static trial was subsequently recorded in Vicon Nexus to register the marker set and compute each participant’s maximum reach.

\subsection{Kinematic Processing}
\subsubsection{Marker-based Processing}
The origin of the local torso coordinate system was defined as the midpoint between the sternal notch and T1, with the vertical vector $\mathbf{V}$ extending from T1 to T8. A placeholder vector in the Anterior–Posterior (AP) direction, $\mathbf{AP}_{\text{temp}}$, was defined from T8 to the sternal notch. The Medial–Lateral (ML) vector, $\mathbf{ML}$, was computed as the cross product of $\mathbf{AP}_{\text{temp}}$ and $\mathbf{V}$: $\mathbf{ML} = \mathbf{AP}_{\text{temp}} \times \mathbf{V}$. The final AP vector was computed as $\mathbf{AP} = \mathbf{V} \times \mathbf{ML}$, forming a right-handed, anatomical torso coordinate system. The end effector at the wrist was defined as the midpoint between the medial and lateral wrist markers. This differs from previous UERW work, which used a marker on the second digit as the end effector \cite{han_reachable_2015, richardson_reachable_2022}. Targets were defined as the maximum distance of the wrist from the origin of the local torso coordinate system. The analysis was performed across all six octants: Superior Anterior Ipsilateral (Sup. Ant. Ipsil.), Superior Anterior Contralateral (Sup. Ant. Contra.), Inferior Anterior Contralateral (Inf. Ant. Contra.), Inferior Anterior Ipsilateral (Inf. Ant. Ipsil.), Inferior Posterior Ipsilateral (Inf. Post. Ipsil.), and Superior Posterior Ipsilateral (Sup. Post. Ipsil.).

\subsubsection{Markerless Processing}
Video data from both camera configurations were first processed using PosePipe~\cite{cottonPosepipe}. We extracted 2D and 3D keypoints using MeTRAbs-ACAE, which outputs a superset of anatomical landmarks. From this set, we utilized the 87 MoVi keypoints, which provide robust coverage across anatomical landmarks relevant to biomechanical reconstruction \cite{sarandiMeTRAbs2021, BLM_Movi,cotton2024differentiablebiomechanicsunlocksopportunities}. This setup makes this system a purely monocular video based system,  that is: the biomechanical reconstructions from the monocular camera are solely based on the 2-D and 3-D data from a single video camera. Person detection and identity association were performed using mmdetDeepSort \cite{mmdetection}.

\textbf{Biomechanical Reconstruction.}
Following Peiffer et al. Portable Biomechanics Laboratory, we fit a whole-body biomechanical model in MuJoCo to the keypoints extracted from videos \cite{peiffer2025portable}. We express the forward kinematic mapping of the model as:
\begin{equation}
    x = M(\theta, \beta)
    \label{eq:motion_model}
\end{equation}
where $\theta \in \mathbb{R}^{40}$ are joint angles, $\beta \in \mathbb{R}^{8 + 87 \times 3}$ contains eight body-scaling parameters and the keypoint offsets, and $x \in \mathbb{R}^{87 \times 3}$ and gives the resulting keypoint positions. MuJoCo’s GPU-accelerated engine enables parallelized forward-kinematic evaluations. The kinematic trajectory for each trial is represented as a learned implicit function $f_{\phi}$, implemented as a multilayer perceptron with sinusoidal temporal encoding~\cite{Vaswani_Attention_2017}:
\begin{equation}
    f_{\phi}: t \mapsto \theta(t)
\end{equation}
Rotational outputs are passed through a $\tanh$ nonlinearity and linearly rescaled to biomechanical joint limits before being evaluated by the forward kinematic model in Equation \ref{eq:motion_model}. Each trial is associated with a dedicated implicit function, and we jointly optimize all trial-specific parameters $\{\phi_i\}_{i=0}^{N-1}$ together with the shared scaling and offset parameters $\beta$, following the bilevel-optimization approach of~\cite{werling2022rapid}. During each optimization step, 300 time samples per trial are evaluated to produce predicted joint angles $\hat{\theta}(t)$, followed by GPU-accelerated forward kinematic evaluations $\hat{x} = M(\hat{\theta}(t), \hat{\beta})$.
The 3D loss is defined as:
\begin{equation}
    L_{\mathrm{3D}} =
    \frac{1}{J} \sum_{j \in J} c(j) \,
    g\!\left(
        \left\|
            \hat{x}_n - x_c
        \right\|_2
    \right),
\end{equation}
where $\hat{x}_n$ are the 3D keypoints measured from the model at each optimization step, $x_c$ are the 3D keypoints extracted from the video, $c(j)$ is the confidence score of each keypoint, and $g(\cdot)$ is a Huber loss (quadratic within 10~cm).

Using the calibrated camera intrinsics and projection model $\Pi$, the 2D loss compares projected model keypoints with detected 2D keypoints $u$:
\begin{equation}
    L_{\mathrm{2D}} =
    \frac{1}{J} \sum_{j \in J} c(j) \,
    g\!\left(
        \left\|
            \Pi(\hat{x}_n) - u
        \right\|_2
    \right),
\end{equation}
with a 5-pixel quadratic region for the Huber loss.

Although the original pipeline was designed for use with mobile devices with an inertial loss term, we have omitted this as our cameras in this work were static. Therefore, the full objective is:
\begin{equation}
    L = \lambda_1 L_{\mathrm{3D}}
      + \lambda_2 L_{\mathrm{2D}},
\end{equation}
with $(\lambda_1, \lambda_2) = (1, 10^{-1})$. Training used Equinox, JAX, and Adam optimization with a learning-rate decay from $10^{-3}$ to $10^{-6}$ and weight decay of $10^{-5}$ \cite{kidger_equinox_2021, deepmind2020jax}. Model training was completed using JAX 0.6.0 on a workstation with 64 GB RAM and an NVIDIA RTX A6000 GPU (16 GB VRAM) \cite{peiffer2025portable, cotton2024differentiablebiomechanicsunlocksopportunities}. 

All assessments of participant movement were performed post-hoc. Coordinate systems derived from the MMC data were aligned with the marker-based coordinate systems, as the MoVi keypoint locations closely matched the anatomical positions of the marker-based system (Fig.~\ref{fig:methods_figure}). The origin of the MMC coordinate system was defined as the midpoint between the “clavicle” and “backneck” keypoints, corresponding approximately to the sternal notch and T1 markers in the marker-based system. The vertical vector $\mathbf{V}$ extended from the “upper back” keypoint (approximately T8) to the “back of neck” keypoint (approximately T1). The placeholder AP vector, $\mathbf{AP}_{\text{temp}}$, was defined from the “upper back” to the “clavicle.” The ML vector was calculated as $\mathbf{ML} = \mathbf{AP}_{\text{temp}} \times \mathbf{V}$, and the final AP vector was $\mathbf{AP} = \mathbf{V} \times \mathbf{ML}$, forming a right-handed anatomical coordinate system. The calculation of maximum reach distance was identical to that used in the marker-based system.

\subsection{Statistical Analysis}
We assessed the performance of monocular MMC throughout the UE workspace using three measurements: (1) peak reach distance differences between marker-based and monocular MMC for each camera configuration; (2) the validity of the percentage of workspace reached (the primary clinical measure of the UERW task) derived from monocular MMC; and (3) the agreement between the marker-based and monocular MMC systems in each octant at every time point.

First, we assessed differences in peak reach distance, measured as the distance from the origin of the torso coordinate system to the wrist in a single frame. This value determines the location of the virtual targets for post-hoc analysis. We then computed the percentage of workspace reached in each octant for the marker-based system and both MMC configurations.

Second, to assess statistical differences between systems and octants regarding the reachable workspace accessed, we used a two-way repeated-measures Analysis of Variance (ANOVA). Greenhouse–Geisser corrections were applied when sphericity was violated, and Bonferroni corrections were used for post-hoc comparisons. The within-subjects factor “system” had three levels: marker-based, frontal, and offset. The within-subjects factor “octant” had six levels: Superior Anterior Ipsilateral (Sup. Ant. Ipsil.), Superior Anterior Contralateral (Sup. Ant. Con.), Superior Posterior Ipsilateral (Sup. Post. Ipsil.), Inferior Anterior Ipsilateral (Inf. Ant. Ipsil.), Inferior Anterior Contralateral (Inf. Ant. Con.), and Inferior Posterior Ipsilateral (Inf. Post. Ipsil.). The dependent variable was the percentage of workspace reached in each octant. The alpha level was set at $\alpha = 0.05$, and all analyses were conducted in Python using the Pingouin package (v0.5.5).

Finally, we assessed whether the marker-based and MMC systems agreed on which octant the end effector occupied at each time point. The rate of agreement was calculated as the percentage of time points where both systems assigned the wrist to the same octant:
$$
\text{Agreement Rate} = \frac{\text{Total Agreements}}{\text{Total Time Points in Octant}} \times 100\%.
$$
When disagreement occurred, the direction of disagreement was identified, and a directional error rate was computed as:
$$
\text{Directional Error Rate} = \frac{\text{Total Disagreements in a Direction}}{\text{Total Time Points in Octant}} \times 100\%.
$$
For example, if the marker-based system measured the end effector in the Sup. Post. Ipsil. octant while the monocular MMC system placed it in the Sup. Ant. Ipsil. octant at the same time point, that frame would be recorded as a disagreement for the Sup. Post. Ipsil. octant. Consequently, the error direction would be identified as the AP direction for the calculation of the directional error rate.

\section{Results}
The differences in peak reach between the marker-based and MMC systems were minimal. Specifically, the differences in estimated reach distances were \(0.04 \pm 0.03\,\text{m}\) for both the offset and frontal camera configurations.

\begin{figure}[ht]
    \centering
    \includegraphics[width=\linewidth]{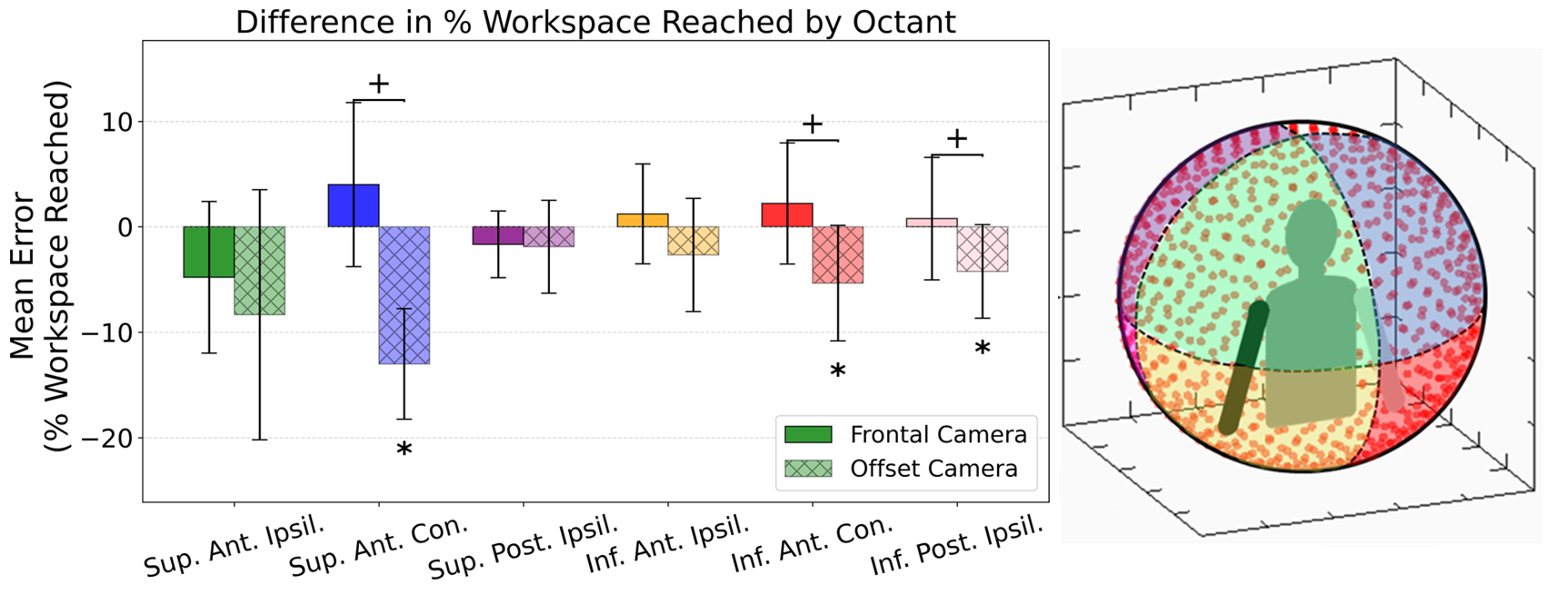}
    \caption{Differences in percentage of reachable workspace (\% Reachspace) for each octant between the marker-based system and both MMC orientations. A difference of 0 indicates perfect agreement between the MMC and the marker-based system. An * indicates a statistically significant difference in the number of targets reached between an MMC orientation and the marker-based workspace (based on pairwise comparisons with Bonferroni correction). A + indicates a statistically significant difference between the frontal and offset orientations in percentage of workspace reached.}
    \label{fig:mae_mmc}
\end{figure}

\begin{table}[h]
\centering

\label{tab:workspace_data}
\begin{tabular}{@{}llc@{}}
\toprule
\textbf{Octant} & \textbf{System} & \textbf{Mean $\pm$ SD (\%)} \\ \midrule

Sup. Ant. Ipsil.  & Marker-Based & 86.44 $\pm$ 8.63 \\ 
 & Frontal  & 81.67 $\pm$ 9.27 \\
 & Offset  & 78.11 $\pm$ 11.73 \\ \midrule

Sup. Ant. Contra. & Marker-Based & 25.89 $\pm$ 11.52 \\ 
& Frontal  & 29.89 $\pm$ 9.55 \\
 & Offset  & 12.89 $\pm$ 8.04 \\  \midrule

Sup. Post. Ipsil.& Marker-Based & 11.78 $\pm$ 8.15 \\
& Frontal  & 10.11 $\pm$ 9.68 \\
& Offset  & 9.89 $\pm$ 10.36 \\ \midrule

Inf. Ant. Ipsil.  & Marker-Based & 82.78 $\pm$ 6.91 \\
 & Frontal  & 84.00 $\pm$ 5.74 \\
 & Offset & 80.11 $\pm$ 7.24 \\ \midrule

Inf. Ant. Contra.  & Marker-Based & 15.22 $\pm$ 9.77 \\
& Frontal & 17.44 $\pm$ 6.82 \\
& Offset & 9.89 $\pm$ 6.51 \\  \midrule

Inf. Post. Ipsil. & Marker-Based & 23.11 $\pm$ 11.06 \\
& Frontal  & 23.89 $\pm$ 13.44 \\
& Offset & 18.89 $\pm$ 10.75 \\
\bottomrule 

\end{tabular}
\caption{Comparison of the Total Percent Workspace Reached in Each Octant for Each Motion Capture System.}
\label{tab:total_workspace_reached}
\end{table}

The total workspace reached by each system is presented in (Table \ref{tab:total_workspace_reached}).   in Differences in the percentage of workspace reached are presented in Fig.~\ref{fig:mae_mmc}.  A two-way repeated-measures ANOVA on the percent workspace reached in each octant revealed significant main effects of \textit{system} \((F(2, 16) = 43.87,\, p < 0.001,\, \eta^2_G = 0.090)\) and \textit{octant} \((F(5, 40) = 223.84,\, p < 0.001,\, \eta^2_G = 0.924)\), as well as a significant \textit{system}~$\times$~\textit{octant} interaction \((F(10, 80) = 4.80,\, p = 0.022,\, \eta^2_G = 0.066)\). Bonferroni-corrected post hoc comparisons showed that the offset camera differed significantly in percentage of workspace reached from both the frontal camera \((t(8) = -6.91,\, p < 0.001,\, g = -0.92)\) and the marker-based system \((t(8) = -5.95,\, p = 0.001,\, g = -0.68)\). In contrast, there was no significant difference between the frontal camera and marker-based systems in \((t(8) = 2.16,\, p = 0.189,\, g = 0.24)\).  Pairwise comparisons of the percentage of reachable workspace accessed revealed several significant differences across the system configurations for specific octants (Table \ref{tab:pairwise_octants_combined}). Crucially, when comparing the Frontal camera system against the Marker-based system, no statistically significant differences in the percentage of reachable workspace were observed across any of the six measured octants (all $p > 0.303$). The effect sizes for these frontal/marker comparisons were consistently small ($\eta^2_p$ ranging from $0.068$ to $0.300$).

In contrast, the offset camera system demonstrated a statistically significant difference compared to the Marker-based system only in the Superior Anterior Contralateral octant ($p < 0.001, \eta^2_p = 0.854$), indicating a large effect. Furthermore, significant differences were found between the two MMC systems (offset/frontal) in the Inferior Anterior Contralateral ($p < 0.001, \eta^2_p = 0.892$) and Superior Anterior Contralateral ($p < 0.001, \eta^2_p = 0.886$) octants, suggesting a substantial directional disparity between the two monocular configurations in the contralateral workspace.

\begin{table}[ht]
\footnotesize
\centering
\begin{tabular}{@{}lccc@{}}
\toprule
\textbf{Octant} & \textbf{Offset/Marker} & \textbf{Frontal/Marker} & \textbf{Offset/Frontal} \\
                & \textbf{p / $\eta^2_p$} & \textbf{p / $\eta^2_p$} & \textbf{p / $\eta^2_p$} \\
\midrule
Inf. Ant. Contra.  & 0.063 / 0.507 & 0.689 / 0.175 & 0.000 / 0.892 \\
Inf. Ant. Ipsil.   & 0.522 / 0.218 & 0.999 / 0.070 & 0.215 / 0.350 \\
Inf. Post. Ipsil.  & 0.075 / 0.486 & 0.999 / 0.068 & 0.056 / 0.520 \\
Sup. Ant. Contra.  & 0.000 / 0.854 & 0.320 / 0.292 & 0.000 / 0.886 \\
Sup. Ant. Ipsil.   & 0.236 / 0.337 & 0.303 / 0.300 & 0.406 / 0.256 \\
Sup. Post. Ipsil.  & 0.757 / 0.160 & 0.551 / 0.209 & 0.999 / 0.002 \\
\bottomrule
\end{tabular}
\caption{Pairwise comparisons of reachable workspace percentages between system configurations for each octant. The table presents the statistical significance ($p$-value) and effect size ($\eta^2_p$) for each comparison.} 
\label{tab:pairwise_octants_combined}
\end{table}

Bland–Altman analysis revealed minimal differences in percentage of workspace reached between the marker-based and frontal camera configurations (Table~\ref{tab:bland_altman_by_octant}) and larger differences for the offset camera configuration. For the frontal camera, mean differences ranged from \(-4.44\%\) to \(4.56\%\). The Sup. Ant. Contra. octant demonstrated the largest positive bias (\(4.56\%\)), whereas the Sup. Ant. Ipsil. octant showed the largest negative bias (\(-4.44\%\)). The 95\% Confidence Intervals (CIs) for individual octants with the frontal camera spanned from \(-18.55\%\) to \(19.29\%\) (Table~\ref{tab:bland_altman_by_octant}).

The offset camera configuration exhibited greater underestimation of reachable workspace compared to the marker-based system, with an overall mean difference across all octants of \(-5.59\%\) (95\% CI: \(-19.90\%\) to \(8.71\%\)). Octant-specific mean differences ranged from \(-19.90\%\) to \(-1.78\%\), with CIs spanning \(-31.34\%\) to \(15.34\%\). The Sup. Ant. Contra. octant showed the largest negative bias (\(-12.44\%\)). There were no positive differences for the offset camera; on average, this configuration underestimated the percentage of workspace reached compared to the marker-based system (Table~\ref{tab:bland_altman_by_octant}).

\begin{table}[ht]
\centering

\begin{tabular}{@{}lccc@{}}
\toprule
\textbf{Octant} & \textbf{Camera} & \makecell{\textbf{Mean Difference} \\ \textbf{(\% Difference)}} & \makecell{\textbf{95\% Confidence} \\ \textbf{Interval}} \\
\midrule
\textbf{All Octants} & Frontal  & -0.70 & -11.70 -  12.90 \\

&  Offset & -5.59 & -19.90 -  8.71 \\
\midrule
Sup. Ant. Ipsil. & Frontal  & -4.78 & -18.87 -  9.31 \\
& Offset  & -8.33 & -31.57 -  14.90 \\
\midrule
Sup. Ant. Contra. & Frontal  & 4.00 & -11.25 -  19.25 \\
& Offset  & -13.00 & -23.28 -  -2.72 \\
\midrule
Sup. Post. Ipsil. & Frontal  & -1.67& -7.86 - 4.53 \\
& Offset  & -1.89 & -10.51 - 6.74\\ 
\midrule
Inf. Ant. Ipsil. & Frontal  & 1.22 & -8.06 -  10.51 \\
& Offset  & -2.67 & -13.18 -  7.84 \\
\midrule
Inf. Ant. Contra. & Frontal  & 2.22 & -9.03 -  13.47 \\
& Offset  & -5.33 & -16.07-  5.40 \\
\midrule
Inf. Post. Ipsil. & Frontal  & 0.78 & -10.60 -  12.15 \\
& Offset  & -4.22  & -12.92 -   4.48 \\
\bottomrule
\end{tabular}
\caption{Bland–Altman statistics comparing percentage of reachable workspace between the marker-based motion capture and two MMC configurations. Mean difference is calculated as (MMC \% Reachable Workspace) – (marker-based \% Reachable Workspace).}
\label{tab:bland_altman_by_octant}
\end{table}

\begin{figure*}[ht]
    \centering
    \includegraphics[width=\textwidth]{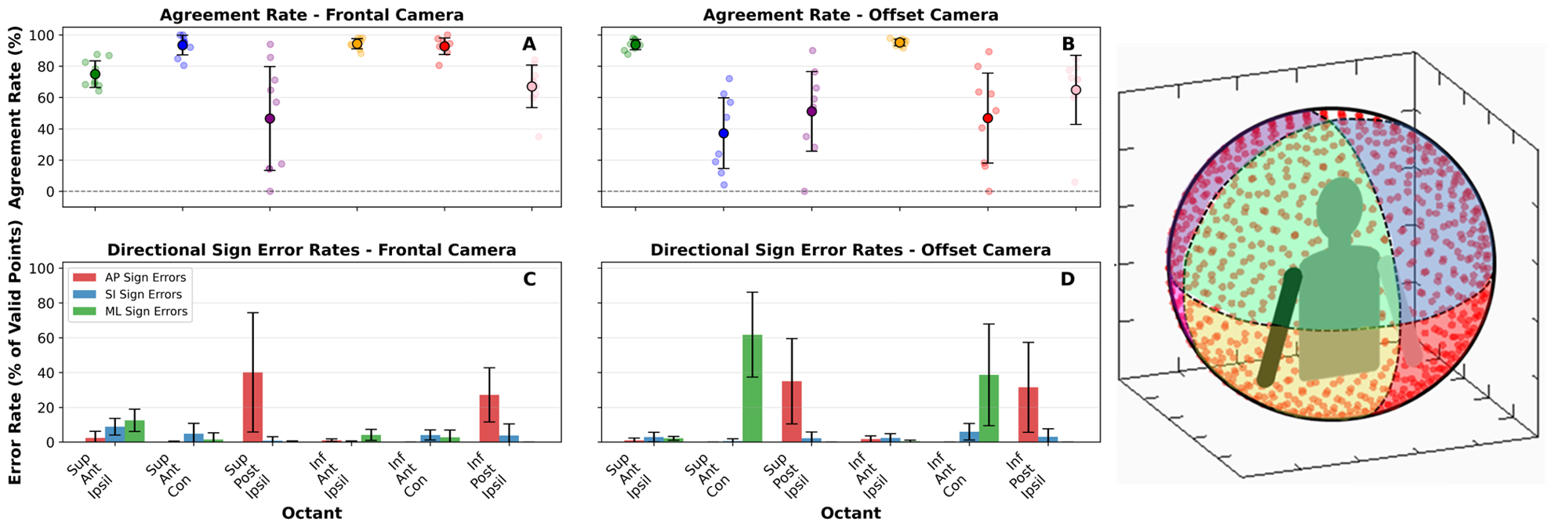}
    \caption{Agreement rate between the marker-based system and both MMC orientations (frontal and offset) across all octants (Panels A \& B). Agreement is expressed as a percentage, with 100\% indicating perfect alignment in octant identification. Panels C and D illustrate directional error rates (Anterior–Posterior, Superior–Inferior, and Medial–Lateral) when systems were not in agreement. Error rates are shown as a percentage of total possible agreement, with 0\% representing no directional error.}
    \label{fig:Agreement and error rate}
\end{figure*}

Agreement rates and directional error rates varied across octants and between the frontal and offset camera configurations (Fig.~\ref{fig:Agreement and error rate}). For the Sup. Ant. Ipsil. octant, the offset camera demonstrated higher agreement with the marker-based system (\(93.83 \pm 3.49\%\)) compared to the frontal camera (\(74.97 \pm 8.99\%\)), with generally lower error rates across the Anterior/Posterior, Superior/Inferior, and Medial/Lateral directions. Conversely, for the Sup. Ant. Contra. octant, the frontal camera achieved higher agreement (\(93.48 \pm 6.67\%\)) with minimal errors, whereas the offset camera showed lower agreement (\(37.21 \pm 23.95\%\)) with elevated ML error (\(61.77 \pm 24.39\%\)). In the Sup. Post. Ipsil. octant, both cameras exhibited lower agreement (frontal: \(46.58 \pm 35.17\%\); offset: \(51.14 \pm 27.03\%\)) with high AP directional errors (frontal: \(40.11 \pm 34.28\%\); offset: \(34.96 \pm 24.53\%\)). For the Inf. Ant. Ipsil. octant, both configurations achieved high agreement (frontal: \(94.32 \pm 3.44\%\); offset: \(95.21 \pm 2.21\%\)) with consistently low directional errors. In the Inf. Ant. Contra. octant, the frontal camera demonstrated higher agreement (\(92.79 \pm 5.62\%\)) than the offset camera (\(46.86 \pm 30.49\%\)), though the offset camera showed higher ML error (\(38.65 \pm 29.24\%\)). Finally, in the Inf. Post. Ipsil. octant, both cameras demonstrated moderate agreement (frontal: \(67.11 \pm 14.43\%\); offset: \(64.89 \pm 23.45\%\)) with elevated AP directional errors (frontal: \(27.16 \pm 15.60\%\); offset: \(31.48 \pm 25.81\%\)).

No statistically significant differences in the percentage of workspace reached were observed between the frontal-plane MMC reconstructions and the marker-based reference. The offset camera configuration, however, differed significantly from the marker-based system and from the frontal configuration in the contralateral anterior octants (Sup. Ant. Contra. and Inf. Ant. Contra.; Table  \ref{tab:pairwise_octants_combined}). 

Agreement analysis (Fig.~\ref{fig:Agreement and error rate}) showed high agreement ($>90\%$) between the frontal camera and the marker-based system in four of six octants, primarily in anterior regions. Reduced agreement and larger directional errors were observed in posterior-facing octants for both systems, predominantly in the anterior–posterior (AP) direction.

\section{Discussion}
This study demonstrates that a single-camera (monocular) MMC approach can accurately quantify the percentage of targets reached during the UERW task in healthy adults. The UERW measurements obtained using the frontal-camera system were comparable to those derived from a traditional marker-based motion capture system, supporting the feasibility of monocular MMC for multi-planar, clinically relevant assessments. To our knowledge, this is the first demonstration that a monocular system can achieve performance similar to a marker-based reference for clinical evaluation of upper extremity mobility, whereas previous work has only validated such approaches using multi-camera systems and the measurement accuracies of the systems and not the clinical outputs \cite{unger_differentiable_2025, hansen_validation_2024}. Other monocular approaches such as \cite {Lagomarsino_2025, panattoni2024hand, Tang_2025} are limited by a lack of anatomically accurate model for clinical decision making, but have been nonetheless shown to be valid tools for the assessment of UE tasks with monocular cameras.  By reducing technical complexity, equipment requirements, and setup time, monocular MMC offers a practical solution for implementing quantitative UE mobility assessments in small clinical teams or point-of-care settings, potentially increasing accessibility and adoption of objective movement analysis in routine practice. 

An important consideration in this analysis is the definition of peak reach distance, which was computed post hoc from both the MMC and marker-based reconstructions. This parameter determines the virtual target locations for each participant for the post-hoc analysis and differs slightly from the VR target placement that is seen by the participants in the headset. We observed small differences between systems in peak reach distance ($0.04 \pm 0.03$~m for both orientations), which are likely attributable to systematic biases in keypoint localization.  Because the detected keypoints are not rigidly attached to anatomical landmarks, small shifts in detection can propagate through the reconstruction pipeline. Nevertheless, given the minimal differences in peak reach distance, subsequent differences in UERW were primarily driven by camera orientation.

Errors in both systems were driven largely by the inherent difficulty of measuring motions that were out of line with the view of the monocular video. The frontal camera demonstrated its poorest agreement in posterior octants, where depth ambiguity produced elevated AP-direction errors (Fig.~\ref{fig:Agreement and error rate}). The offset camera showed a comparable trend, with substantial AP-direction errors in posterior octants and ML-direction errors in contralateral octants. These patterns corroborate previous research showing that measurements are most accurate in the anatomical plane that aligns with the camera view \cite{peiffer2025portable}. The concentration of error in the posterior and contralateral octants suggests a sensitivity to perspective projection constraints. Specifically, when movement vectors align with the camera’s optical axis, the resulting foreshortening reduces the signal-to-noise ratio of 2D keypoint displacement. This creates a reliance on the biomechanical model’s depth-estimation heuristics, which, as evidenced by our AP-direction error rates, struggles to resolve depth ambiguity in the absence of binocular or multi-view parity.

Overall, however, the frontal camera orientation showed no statistically significant difference from the marker-based assessment and may serve as a viable clinical solution for the assessment of UERW, particularly for the anterior workspace. Deviations were typically less than five targets per octant, well below deficits observed in upper extremity clinical populations, where differences between affected and unaffected limbs have been shown to be up to 35 targets per octant reached between affected and unaffected limbs in pediatric patients with Brachial Plexus Palsy with the exceptions of the Inf. Ant. Ipsil. and Contra. octants that had differences of less than 10 targets per octant \cite{ richardson_evaluation_2024}. With our cohort we are well below the differences observed in clinical populations. For reaches into the superior anterior octants, which require shoulder elevation, a function commonly impaired in patient groups such as stroke survivors and children with brachial plexus palsy, the frontal camera configuration performed well, showing no significant difference from marker-based motion capture in either of the four anterior octants \cite{russo2014scapulothoracic, fitoussi2009upper, eismann2015glenohumeral}. This is consistent with prior findings for tasks conducted within the same workspace \cite{unger_differentiable_2025, hansen_validation_2024}.

In contrast, the offset camera orientation significantly underestimated the number of targets reached in several key regions, including the superior-anterior contralateral, inferior-anterior contralateral, and inferior-posterior ipsilateral octants (Fig.~\ref{fig:mae_mmc}). These discrepancies underscore a fundamental trade-off in monocular motion capture between perspective-dependent depth resolution and anatomical occlusion.While our findings support the general feasibility of single-camera systems, they highlight that no single viewpoint is currently optimal for tasks requiring dynamic, multi-planar motion that aligns with the camera's optical axis. For example, the offset camera achieved less than 40\% agreement in the superior-anterior contralateral octant, exhibiting large ML errors. Mechanistically, this poor performance is triggered by cross-body movements where the torso or the reaching arm itself physically obstructs the camera’s line of sight to the hand and wrist.During these periods of occlusion, the MMC algorithm must "infer" distal joint positions, leading to inaccurate depth estimations. This stands in stark contrast to the Inf. Ant. Ipsil. octant, where both camera orientations achieved high agreement rates ($>95\%$). In this zone, the movement occurs largely within the camera's optimal plane of view, minimizing both occlusion and projection foreshortening. Consequently, the frontal camera showed minimal bias (1.22\% mean difference), whereas the offset camera’s persistent underestimation in other zones reflects the current algorithmic difficulty in resolving depth when the limb’s movement vector parallels the camera lens.

While this study demonstrates the feasibility of monocular MMC, several constraints must be addressed to ensure clinical utility. First, the generalization of these findings is limited by the homogeneity of the study cohort, which consisted entirely of healthy adults. Clinical populations, such as stroke survivors or children with brachial plexus palsy, often exhibit atypical movement kinematics, including tremors, slower velocities, or compensatory trunk movements. It remains to be determined if the current detection algorithms maintain their level of precision when faced with these neuromotor pathologies, as the "negligible" errors observed here fewer than five targets per octant—may have greater functional implications in patients. For instance, in severe brachial plexus injuries where a patient may only reach the inner margins of the workspace, a small depth-estimation error could be the deciding factor in whether a target is classified as "reached." 

Furthermore, the current analysis was conducted post-hoc, which limits its immediate utility at the point of care. For monocular MMC to be adopted into routine clinical workflows, the transition from data capture to actionable reporting must be streamlined. These results should be interpreted with caution, particularly for localized analysis, given that average percent errors reached as high as 15.45\% (Frontal) and 50.21\% (Offset) in contralateral octants. Such high regional variability indicates that while the system may be suitable for assessing total reachable workspace, it lacks the consistency required for precise measurement of specific superior or contralateral octants.Future iterations should focus on integrating real-time feedback and leveraging mobile platforms to provide immediate biofeedback during the assessment. Such improvements would reduce the technical overhead even further, making objective motion analysis accessible in both busy clinical environments and home-based rehabilitation settings. Despite these challenges, the single-camera approach significantly reduces the equipment requirements and setup time associated with traditional marker-based systems. By addressing these spatial and algorithmic limitations, monocular MMC offers a scalable solution for expanding the reach of objective movement analysis in diverse clinical settings. 

To our knowledge, this is the first study to evaluate the feasibility of a monocular, video-based system for the analysis of UERW in a clinical context. Our results demonstrate that a frontal camera orientation provides UERW measurements comparable to those of gold-standard marker-based motion capture, particularly within the anterior octants. By establishing a single-camera setup as a reliable alternative to complex, multi-camera arrays, this approach significantly lowers the barrier to entry for objective movement analysis. Ultimately, this work provides a foundation for the automation of a common clinical assessment, offering a practical path toward integrating quantitative upper-extremity metrics into routine point-of-care workflows.

\bibliographystyle{plain}
\bibliography{References} 

\begin{thebibliography}{10}

\bibitem{abzug2010current}
J.M. Abzug and S.H. Kozin.
\newblock Current concepts: neonatal brachial plexus palsy.
\newblock {\em Orthopedics}, 33(6):430--435, 2010.

\bibitem{mmdetection}
K.~Chen, J.~Wang, J.~Pang, Y.~Cao, Y.~Xiong, X.~Li, S.~Sun, W.~Feng, Z.~Liu, J.~Xu, Z.~Zhang, D.~Cheng, C.~Zhu, T.~Cheng, Q.~Zhao, B.~Li, X.~Lu, R.~Zhu, Y.~Wu, J.~Dai, J.~Wang, J.~Shi, W.~Ouyang, C.~C. Loy, and D.~Lin.
\newblock {MMDetection}: Open mmlab detection toolbox and benchmark.
\newblock {\em arXiv preprint arXiv:1906.07155}, 2019.

\bibitem{cottonPosepipe}
R.~J. Cotton.
\newblock Posepipe: Open-source human pose estimation pipeline for rehabilitation research.
\newblock {\em Archives of Physical Medicine and Rehabilitation}, 103(12):e161--e162, 2022.

\bibitem{cotton2024differentiablebiomechanicsunlocksopportunities}
R.~J. Cotton.
\newblock Differentiable biomechanics unlocks opportunities for markerless motion capture, 2024.

\bibitem{CottonGaitTransformer}
R.~James Cotton, Emoonah McClerklin, Anthony Cimorelli, Ankit Patel, and Tasos Karakostas.
\newblock Transforming gait: Video-based spatiotemporal gait analysis.
\newblock In {\em 2022 44th Annual International Conference of the IEEE Engineering in Medicine \& Biology Society (EMBC)}, pages 115--120, 2022.

\bibitem{Davids2006SHUEE}
Jon~R. Davids, Laura~C. Peace, Lisa~V. Wagner, Mary~Ann Gidewall, Dawn~W. Blackhurst, and W.~Matthew Roberson.
\newblock Validation of the shriners hospital for children upper extremity evaluation (shuee) for children with hemiplegic cerebral palsy.
\newblock {\em The Journal of Bone \& Joint Surgery}, 88(2):326--333, Feb 2006.

\bibitem{de_bie_longitudinal_2017}
E.~de~Bie, B.~Oskarsson, N.~C. Joyce, A.~Nicorici, G.~Kurillo, and J.~J. Han.
\newblock Longitudinal evaluation of upper extremity reachable workspace in als by kinect sensor.
\newblock {\em Amyotrophic Lateral Sclerosis and Frontotemporal Degeneration}, 18(1-2):17--23, February 2017.

\bibitem{deepmind2020jax}
{DeepMind}, Igor Babuschkin, Kate Baumli, Alison Bell, Surya Bhupatiraju, Jake Bruce, Peter Buchlovsky, David Budden, Trevor Cai, Aidan Clark, Ivo Danihelka, Antoine Dedieu, Claudio Fantacci, Jonathan Godwin, Chris Jones, Ross Hemsley, Tom Hennigan, Matteo Hessel, Shaobo Hou, Steven Kapturowski, Thomas Keck, Iurii Kemaev, Michael King, Markus Kunesch, Lena Martens, Hamza Merzic, Vladimir Mikulik, Tamara Norman, George Papamakarios, John Quan, Roman Ring, Francisco Ruiz, Alvaro Sanchez, Laurent Sartran, Rosalia Schneider, Eren Sezener, Stephen Spencer, Srivatsan Srinivasan, Miloš Stanojevi{ć}c, Wojciech Stokowiec, Luyu Wang, Guangyao Zhou, and Fabio Viola.
\newblock The deepmind jax ecosystem, 2020.

\bibitem{eismann2015glenohumeral}
E.A. Eismann, K.J. Little, T.~Laor, and R.~Cornwall.
\newblock Glenohumeral abduction contracture in children with unresolved neonatal brachial plexus palsy.
\newblock {\em J Bone Joint Surg Am}, 97(2):112--118, 2015.

\bibitem{fitoussi2009upper}
F.~Fitoussi, N.~Maurel, A.~Diop, et~al.
\newblock Upper extremity kinematics analysis in obstetrical brachial plexus palsy.
\newblock {\em Orthop Traumatol Surg Res}, 95(5):336--342, 2009.

\bibitem{BLM_Movi}
S.~Ghorbani, K.~Mahdaviani, A.~Thaler, K.~Kording, D.~J. Cook, G.~Blohm, and N.~F. Troje.
\newblock Movi: A large multi-purpose human motion and video dataset.
\newblock {\em PLOS ONE}, 16(6):1--15, June 2021.

\bibitem{GRAF2025}
A.~Graf, J.~J. Krzak, K.~M. Kruger, J.~Davids, R.~Smith, B.~Steinlein, and A.~Bagley.
\newblock Automated identification of clinically meaningful biomechanical phenotypes in cerebral palsy through multicenter gait data.
\newblock {\em Clinical Biomechanics}, 125:106501, 2025.

\bibitem{guo2025artificial}
Ling Guo, Richard Chang, Jie Wang, Amudha Narayanan, Peisheng Qian, Mei~Chee Leong, Partha~Pratim Kundu, Sriram Senthilkumar, Sai~Chaitanya Garlapati, Elson Ching~Kiat Yong, et~al.
\newblock Artificial intelligence-enhanced 3d gait analysis with a single consumer-grade camera.
\newblock {\em Journal of Biomechanics}, 187:112738, 2025.

\bibitem{han2013validity}
J.~J. Han, G.~Kurillo, R.~T. Abresch, A.~Nicorici, and R.~Bajcsy.
\newblock Validity, reliability, and sensitivity of a 3d vision sensor-based upper extremity reachable workspace evaluation in neuromuscular diseases.
\newblock {\em PLoS Currents}, 5:1--21, 2013.

\bibitem{han_reachable_2015}
Jay~J Han, Gregorij Kurillo, Richard~T Abresch, Evan de~Bie, Alina Nicorici, and Ruzena Bajcsy.
\newblock Reachable workspace in facioscapulohumeral muscular dystrophy ({FSHD}) by kinect.
\newblock {\em Muscle \& nerve}, 51(2):168--175, 2015.
\newblock Publisher: Wiley Online Library.

\bibitem{han2015dystrophinopathy}
J.J. Han, G.~Kurillo, R.T. Abresch, et~al.
\newblock Upper extremity 3-dimensional reachable workspace analysis in dystrophinopathy using kinect.
\newblock {\em Muscle Nerve}, 52(3):344--355, 2015.

\bibitem{hansen_validation_2024}
R.~M. Hansen, S.~L. Arena, and R.~M. Queen.
\newblock Validation of upper extremity kinematics using markerless motion capture.
\newblock {\em Biomedical Engineering Advances}, 7:100128, 2024.

\bibitem{kanko_inter-session_2021}
R.~M. Kanko, E.~Laende, W.~S. Selbie, and K.~J. Deluzio.
\newblock Inter-session repeatability of markerless motion capture gait kinematics.
\newblock {\em Journal of Biomechanics}, 121:110422, 2021.

\bibitem{Kephart2020GaitApp}
D.~T. Kephart, S.~R. Laing, A.~Bagley, J.~R. Davids, and V.~A. Kulkarni.
\newblock Gait analysis at your fingertips: Accuracy and reliability of mobile app enhanced observational gait analysis in children with cerebral palsy.
\newblock {\em Journal of the Pediatric Orthopaedic Society of North America}, 2(1):94, 2020.

\bibitem{kidger_equinox_2021}
Patrick Kidger and Cristian Garcia.
\newblock Equinox: neural networks in jax via callable pytrees and filtered transformations, 2021.

\bibitem{klein2024angle}
Luis~C. Klein, Abdelghani~A. Chellal, Vasco Grilo, Jan Braun, Jorge Gonçalves, Maria~F. Pacheco, Filipe~P. Fernandes, Francisco~C. Monteiro, and José Lima.
\newblock Angle assessment for upper limb rehabilitation: A novel light detection and ranging (lidar)-based approach.
\newblock {\em Sensors}, 24(2):530, Jan 2024.

\bibitem{koleini2025biopose}
F.~Koleini, M.~U. Saleem, P.~Wang, H.~Xue, A.~Helmy, and A.~Fenwick.
\newblock Biopose: Biomechanically accurate 3d pose estimation from monocular videos.
\newblock {\em arXiv preprint arXiv:2501.07800}, 2025.

\bibitem{koman2008quantification}
L.A. Koman, R.M.M. Williams, P.J. Evans, et~al.
\newblock Quantification of upper extremity function and range of motion in children with cerebral palsy.
\newblock {\em Dev Med Child Neurol}, 50(12):910--917, 2008.

\bibitem{Lagomarsino_2025}
Beatrice Lagomarsino, Antonino Massone, Francesca Odone, Maura Casadio, and Matteo Moro.
\newblock Video-based markerless assessment of bilateral upper limb motor activity following cervical spinal cord injury.
\newblock {\em Computers in Biology and Medicine}, 196:110908, 2025.

\bibitem{Lovette2024}
M.~Lovette, R.~S. Chafetz, S.~A. Russo, S.~H. Kozin, and D.~A. Zlotolow.
\newblock Shoulder motion overestimated by mallet scores.
\newblock {\em Journal of Pediatric Orthopaedics}, 44(10):e951--e956, November 2024.

\bibitem{nicholson2014evaluating}
K.F. Nicholson, S.A. Russo, S.H. Kozin, et~al.
\newblock Evaluating the acromion marker cluster as a method for measuring scapular orientation in children with brachial plexus birth palsy.
\newblock {\em J Appl Biomech}, 30(1):128--133, 2014.

\bibitem{Olleac2025VideoGait}
R.~Olleac, B.~Centeno, C.~Duffy, M.~Nu{\~n}ez, M.~Crespo, L.~M. Barrios, J.~R. Davids, and V.~A. Kulkarni.
\newblock Vertical video-based gait analysis for assessment of transverse plane motion: Reliability and validity in a neuromuscular population.
\newblock {\em Journal of Pediatric Orthopaedics}, 45(5):294--299, May 2025.

\bibitem{panattoni2024hand}
Alessandro Panattoni, Francesco Betti, Fabio Gualandi, Giovanni Saggio, Gianluca Costante, and Marco Grimaldi.
\newblock Hand tracking for clinical applications: Validation of the google mediapipe framework for finger tapping assessment.
\newblock {\em Biomedical Signal Processing and Control}, 96:106603, 2024.

\bibitem{pantzar2018knee}
Elisa Pantzar-Castilla, Andrea Cereatti, G.~Figari, N.~Valeri, G.~Paolini, U.~Della~Croce, A.~Magnuson, and J.~Riad.
\newblock Knee joint sagittal plane movement in cerebral palsy: a comparative study of 2-dimensional markerless video and 3-dimensional gait analysis.
\newblock {\em Acta Orthopaedica}, 89(6):656--661, Dec 2018.

\bibitem{pearl2014assessing}
M.L. Pearl, F.~Van De~Bunt, M.~Pearl, et~al.
\newblock Assessing shoulder motion in children: age limitations to mallet and abc loops.
\newblock {\em Clin Orthop Relat Res}, 472(2):740--748, 2014.

\bibitem{peiffer2025portable}
J.~D. Peiffer, K.~Shah, I.~Djuraskovic, S.~Anarwala, K.~Abdou, R.~Patel, P.~Jayabalan, B.~Pennicooke, and R.~J. Cotton.
\newblock Portable biomechanics laboratory: Clinically accessible movement analysis from a handheld smartphone.
\newblock {\em arXiv preprint}, 2025.

\bibitem{pierzchlewicz2024platyposecalibratedzeroshotmultihypothesis}
P.~A. Pierzchlewicz, C.~O. da~Silva, R.~J. Cotton, and F.~H. Sinz.
\newblock Platypose: Calibrated zero-shot multi-hypothesis 3d human motion estimation, 2024.

\bibitem{richardson_reachable_2022}
R.~T. Richardson, S.~A. Russo, R.~S. Chafetz, S.~Warshauer, E.~Nice, S.~H. Kozin, D.~A. Zlotolow, and J.~G. Richards.
\newblock Reachable workspace with real-time motion capture feedback to quantify upper extremity function: A study on children with brachial plexus birth injury.
\newblock {\em Journal of Biomechanics}, 132:110939, February 2022.

\bibitem{richardson_evaluation_2024}
R.~T. Richardson, S.~A. Russo, R.~S. Chafetz, S.~Warshauer, E.~Nice, J.~G. Richards, D.~A. Zlotolow, and S.~H. Kozin.
\newblock Evaluation of upper extremity reachable workspace in children with brachial plexus birth injury.
\newblock {\em The Journal of Hand Surgery}, 49(2):141--149, 2024.

\bibitem{russo2014scapulothoracic}
S.~A. Russo, S.~H. Kozin, D.~A. Zlotolow, K.~F. Thomas, R.~L. Hulbert, J.~M. Mattson, K.~M. Rowley, and J.~G. Richards.
\newblock Scapulothoracic and glenohumeral contributions to motion in children with brachial plexus birth palsy.
\newblock {\em J. Shoulder Elbow Surg.}, 23(3):327--338, 2014.

\bibitem{russo2019motion}
S.A. Russo, S.H. Kozin, D.A. Zlotolow, K.F. Nicholson, and J.G. Richards.
\newblock Motion necessary to achieve mallet internal rotation positions in children with brachial plexus birth palsy.
\newblock {\em J Pediatr Orthop}, 39(1):14--21, 2019.

\bibitem{russo2015limited}
S.A. Russo, B.J. Loeffler, D.A. Zlotolow, et~al.
\newblock Limited glenohumeral cross-body adduction in children with brachial plexus birth palsy: a contributor to scapular winging.
\newblock {\em J Pediatr Orthop}, 35(3):240--245, 2015.

\bibitem{russo_motion_2019}
Stephanie~A. Russo, Scott~H. Kozin, Dan~A. Zlotolow, Kristen~F. Nicholson, and James~G. Richards.
\newblock Motion necessary to achieve mallet internal rotation positions in children with brachial plexus birth palsy.
\newblock {\em Journal of Pediatric Orthopaedics}, 39(1):14--21, 2019.

\bibitem{sarandiMeTRAbs2021}
I.~Sárándi, T.~Linder, K.~O. Arras, and B.~Leibe.
\newblock Metrabs: Metric-scale truncation-robust heatmaps for absolute 3d human pose estimation.
\newblock {\em IEEE Transactions on Biometrics, Behavior, and Identity Science}, 3(1):16--30, 2021.

\bibitem{vrraise}
S.~Q.~Y. Tan, Y.~Zhong, B.~Marteau, A.~Hornback, W.~Shi, S.~Warshauer, R.~Courter, S.~Donahue, L.~Lottiers, S.~Russo, R.~T. Richardson, R.~Chafetz, and M.~D. Wang.
\newblock Vr-raise: Virtual reality for reachability assessment using an interactive system environment.
\newblock In {\em 2024 IEEE 3rd International Conference on Intelligent Reality (ICIR)}, pages 1--8, 2024.

\bibitem{Tang_2025}
R.~Tang et~al.
\newblock Using mediapipe to track upper-limb reaching movements after stroke: a proof-of-principle study.
\newblock {\em Journal of NeuroEngineering and Rehabilitation}, 22(1):14, 2025.

\bibitem{uhlrich_opencap_2023}
S.~D. Uhlrich, A.~Falisse, Ł. Kidziński, J.~Muccini, M.~Ko, A.~S. Chaudhari, J.~L. Hicks, and S.~L. Delp.
\newblock {OpenCap}: Human movement dynamics from smartphone videos.
\newblock {\em PLoS Computational Biology}, 19(10):e1011462, 2023.

\bibitem{unger_differentiable_2025}
Tim Unger, Arash~Sal Moslehian, J.D. Peiffer, Johann Ullrich, Roger Gassert, Olivier Lambercy, R.~James Cotton, and Chris~Awai Easthope.
\newblock Differentiable biomechanics for markerless motion capture in upper limb stroke rehabilitation: A comparison with optical motion capture.
\newblock {\em {IEEE} Transactions on Medical Robotics and Bionics}, pages 1--1, 2025.

\bibitem{Vaswani_Attention_2017}
Ashish Vaswani, Noam Shazeer, Niki Parmar, Jakob Uszkoreit, Llion Jones, Aidan~N Gomez, \L~ukasz Kaiser, and Illia Polosukhin.
\newblock Attention is all you need.
\newblock In I.~Guyon, U.~Von Luxburg, S.~Bengio, H.~Wallach, R.~Fergus, S.~Vishwanathan, and R.~Garnett, editors, {\em Advances in Neural Information Processing Systems}, volume~30. Curran Associates, Inc., 2017.

\bibitem{wade_applications_2022}
L~Wade, L~Needham, P~{McGuigan}, and J~Bilzon.
\newblock Applications and limitations of current markerless motion capture methods for clinical gait biomechanics.
\newblock {\em {PeerJ}.}

\bibitem{werling2022rapid}
K.~Werling, M.~Raitor, J.~Stingel, J.~L. Hicks, S.~Collins, S.~L. Delp, and C.~K. Liu.
\newblock Rapid bilevel optimization to concurrently solve musculoskeletal scaling, marker registration, and inverse kinematic problems for human motion reconstruction.
\newblock {\em bioRxiv}, 2022.
\newblock Cold Spring Harbor Laboratory preprint.

\bibitem{wren_comparison_2023}
Tishya~{AL} Wren, Pavel Isakov, and Susan~A Rethlefsen.
\newblock Comparison of kinematics between theia markerless and conventional marker-based gait analysis in clinical patients.
\newblock {\em Gait \& Posture}, 104:9--14, 2023.
\newblock Publisher: Elsevier.

\end{thebibliography}
\end{document}